\documentclass{article}
\usepackage{spconf,amsmath,graphicx,multirow, array}
\usepackage[symbol*]{footmisc}

\title{Learning Mutually Local-global U-nets \\ For High-resolution Retinal Lesion Segmentation in Fundus Images }
\name{Zizheng Yan$^{1}$, Xiaoguang Han$^{1,2,*}$, Changmiao Wang$^{1,4}$, Yuda Qiu$^{1}$, Zixiang Xiong$^{3}$, Shuguang Cui$^{1,2}$}
\address{$^{1}$School of Science and Engineering, The Chinese University of Hong Kong (Shenzhen), China \\
      $^{2}$ Shenzhen Research Institute of Big Data, China  \\
      $^{3}$ Department of Electrical and Computer Engineering, Texas A\&M University, USA \\
      $^{4}$ School of Information Science and Technology, University of Science and Technology of China, China}

\begin{document}
%
\maketitle
\begin{abstract}
Diabetic retinopathy is the most important complication of diabetes. Early diagnosis of retinal lesions helps to avoid visual loss or blindness. Due to high-resolution and small-size lesion regions, applying existing methods, such as U-Nets, to perform segmentation on fundus photography is very challenging. Although downsampling the input images could simplify the problem, it loses detailed information. Conducting patch-level analysis helps reaching fine-scale segmentation yet usually leads to misunderstanding as the lack of context information. In this paper, we propose an efficient network that combines them together, not only being aware of local details but also taking fully use of the context perceptions. This is implemented by integrating the decoder parts of a global-level U-net and a patch-level one. The two streams are jointly optimized, ensuring that they are enhanced mutually. Experimental results demonstrate our new framework significantly outperforms existing patch-based and global-based methods, especially when the lesion regions are scattered and small-scaled.
\end{abstract}
\begin{keywords}
diabetic retinopathy, deep learning, lesion segmentation, local-global U-nets
\end{keywords}
\section{Introduction}
\label{sec:intro}

\footnotetext[1]{Corresponding author}
\footnotetext[3]{This work was funded in part by “The Pearl River Talent Recruitment Program Innovative and Entrepreneurial Teams in 2017” under grant No. 2017ZT07X152 and Shenzhen Fundamental Research Fund under grants No. KQTD2015033114415450 and No. ZDSYS201707251409055.}

Diabetic retinopathy is now a common disease especially among working-age people, also considered as the main cause of blindness. To conduct clinical examination of it, ophthalmologists usually use fundus photography techniques to display the back of the eyeball in a very high-resolution image. On the images, retinal lesions can be apparently visualized. For example, microaneurysms appear as dark spots while hard exudates abnormally display bright regions. Manually carrying out such diagnosis is very non-objective, time-consuming, and highly depends on expertise. It is highly desirable to automate this procedure. To this end, many approaches have been proposed ~\cite{Chudzik2018,amin2016review}, formulating the problem as segmenting those lesion regions, by pixel-wise binary labeling.

Recently, deep Fully Convolutional Neural networks (FCN) ~\cite{Shelhamer2017} has gained much popularity in the field of image segmentation for its ability to learn the most discriminative pixel-wise features. A sequence of convolution and pooling layers are used to form an encoder to covert the input image into feature maps. These feature maps are then decoded to a segmentation mask with another set of deconvolution layers. Based on this architecture, U-net ~\cite{Ronneberger2015} introduces skip concatenation between the encoder and the decoder layers. This improvement reduces the dependence on large samples and results in much better performance. The work stimulates many other variants such as V-net~\cite{Milletari2016} and Segnet~\cite{Badrin2017}. These deep learning architectures are so efficient that they have been commonly used in various applications of medical images segmentation ~\cite{Ronneberger2015,Milletari2016}.

However, the methods above tend to fail in our settings. The significantly high-resolution (usually upto 3500$\times$3500) fundus images with small-size target lesion regions burden the computational resources, and increase the difficulties of learning. By downsampling the input image and then rescaling the output as the final result, those architectures could be adopted yet hard to reach fine segmentation due to loss of information caused by downsampling. Many attempts have been made ~\cite{Lam2018,yang2017lesion,chudzik2018exudate} to avoid such phenomena. Those works split the image into several patches and conduct patch-level segmentation on them respectively. Although these methods preserve detailed information, they usually cause mislabeling of lesion regions and inconsistency across patches, as a result of the poor capture of global contexts.

In this paper, a novel network architecture is proposed to overcome the shortcomings of both global-level and patch-level approaches, by combining them in a unified learning framework. In particular, it consists of two streams: one global stream performs segmentation in a downsampled version that converts the input image into low-resolution label maps; another local stream takes cropped patches as inputs and produces their corresponding segmentation results. These two, each exploiting U-Net as the basic component, are integrated by concatenating the outcome feature maps of the global decoder to the local decoder part. Then, the two steams are jointly optimized. With the whole network well trained, we next conduct segmentation on patches and stitch the outputs together to get our final results. This mechanism benefits the performance in two aspects: 1) the learned context features in the global part are passed to the local stream, reducing ambiguities and correcting errors; 2) the losses in the local stream are passed to global part, enhancing the learning of context features to improve the performance of the local component. In this way, the local and global nets are mutually enhanced. We tested our approach on the public fundus images dataset and conducted segmentation for microaneurysms (MA), soft exudates (SE), hard exudates (EX) and hemorrhages (HE). Our experiments showed that our model significantly outperforms local-only net and global-only net for MA and EX. And in the case of SE and HE, the global-only net outperforms all other variants. We found that the global-only net is more suitable when the lesion regions are compact and large. In contrast, the network proposed will do better when the lesion regions are scattered and of small size.

\section{Method}
\label{sec:format}

The overview of our proposed algorithm is shown in Fig. \ref{fig:net}. Generally, there are three components in our model: GlobalNet, LocalNet and Fusion module. GlobalNet accepts a downsampled version of the image as input and produces a coarse segmentation map with the same size as the input. LocalNet accepts cropped image patches as input and produces segmentation maps in their original resolution. The proposed fusion module is used to crop the feature maps from GlobalNet and concatenate them to the LocalNet so that both global and local information can be captured.

\subsection{Network Architecture}
\label{sec:global}

\textbf{GlobalNet.} In this paper, we adopt U-Net as the backbone of our GlobalNet. The U-Net consists of an encoder and a decoder. The encoder is stacked by conv-bn-relu-conv-bn-relu basic blocks. Along the downsampling path in the encoder, the height and width of the feature map are halved while the number of channels doubles. The decoder architecture exactly follows an inverse encoder, whose feature spatial size doubles while the number of channels halves. In addition, the feature maps with same spatial size in the encoder and the decoder are concatenated.
We adopt U-Net with 6 pooling layers for EX and HE, 4 pooling layers for SE and 3 pooling layers for MA.

\textbf{LocalNet.} We also adopt U-Net as the backbone of our LocalNet. Different from GlobalNet, the inputs of the LocalNet are image patches with smaller spatial size. In our experiments, we adopt U-Net with 3 pooling layers for EX, MA and 6 pooling layers for HE and SE.

\begin{figure}[t!]
	\centering
	\includegraphics[width=9cm]{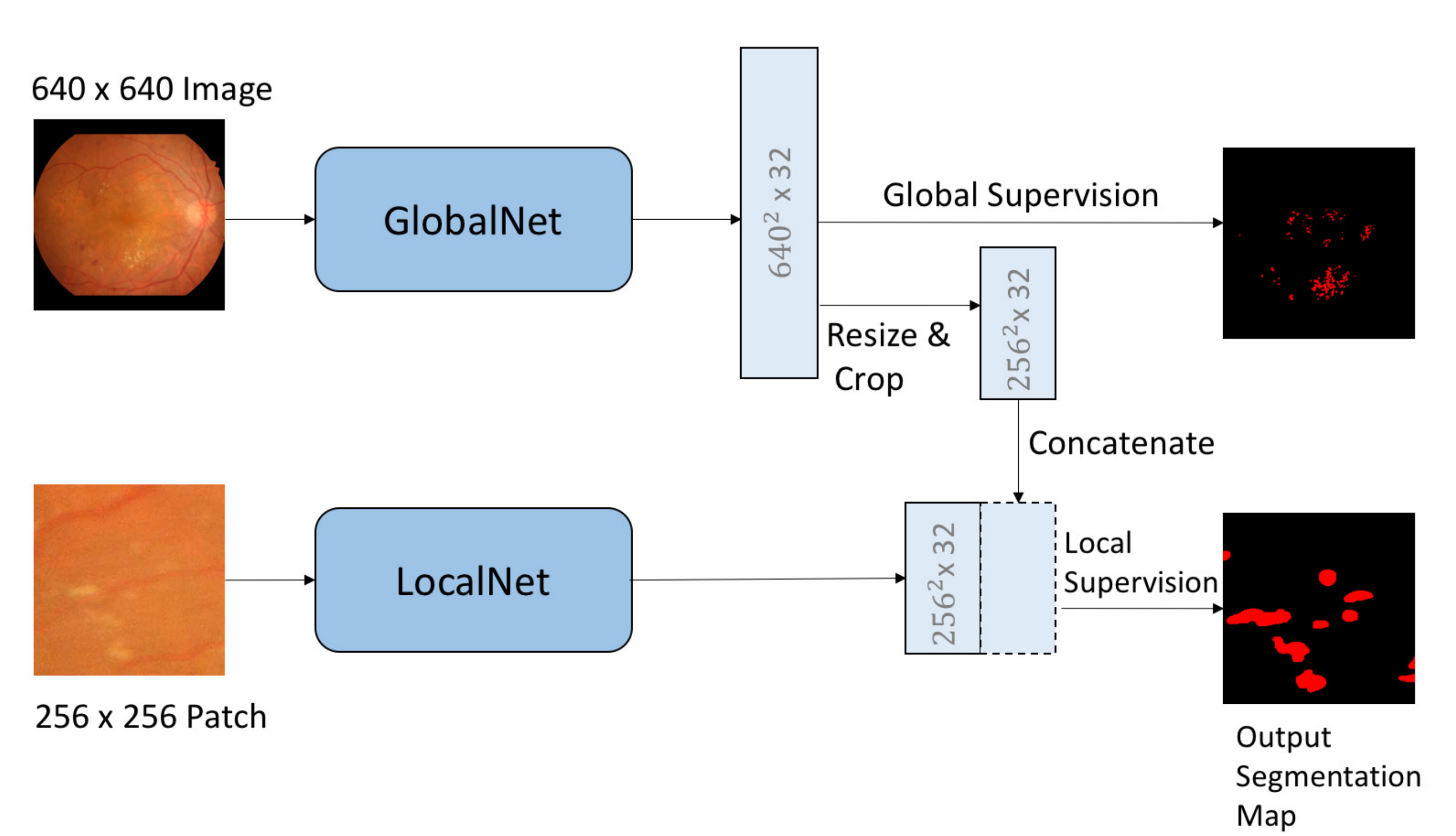}
	\caption{The proposed segmentation network. Both the global supervision and the local supervision are exploited in our training procedure. The output from the LocalNet is used as our final segmentation map when testing.}
	\label{fig:net}
\end{figure}

\textbf{Feature Fusion.} As shown in Fig \ref{fig:net}, the LocalNet and GlobalNet are fused in the end of their decoder component. In particular, the feature map $f_g$ in the end of the global decoder, before outputting the segmentation maps, is firstly took out. Then, it is concatenated to the output feature map $f_l$ in the end of the local decoder, forming a new feature map $f'_l$. As the GlobalNet takes the downsampled original images as inputs while the LocalNet takes cropped patches as inputs, to build correspondences, rescaling and cropping are conducted on $f_g$ before the concatenation. Finally, two 3$\times$3 and one 1$\times$1 convolution layers are exploited to transform $f'_l$ to produce patched segmentation map.

\subsection{Training}
\label{sec:multiscale}

\textbf{Dataset.} The dataset used in this paper is provided by 2018 ISBI grand challenge on diabetic retinopathy segmentation and grading \cite{porwal2018}. We use the dataset of segmentation sub-challenge. This dataset consists of 81 color fundus images with signs of diabetic retinopathy (DR) and other 164 without signs of DR. We only adopt the images with DR, specifically, 81 images of MA, 81 images of hard EX, 80 images of HE, 40 images of SE. Each image with sign of DR may contain more than one abnormality. Generally, the dataset was split to 54 training samples and 27 testing samples by the organizer.

The resolution of original image is 2848$\times$4288 with zero fillings on both sides. We first center crop the image to 2816$\times$3328 to eliminate the zeros fillings. For GlobalNet, we downsample the image to 640$\times$640, while 256$\times$256 patches are cropped uniformly for LocalNet.

Before training the LocalNet and the GlobalNet, the data are augmented by random rotation (with 359 degrees), zooming, flipping and adding random noise. When augmenting the data for the training of the fused net, only rotation with 90, 180, 270 degrees are exploited since the fusion module requires accurate alignment.

\textbf{Loss function.} We keep the supervision of GlobalNet when training our fused network. Thus, the loss is defined as follows:
\begin{equation}
L = \lambda_1 L_{global} + \lambda_2 L_{local} + \lambda_3\phi(\theta)
\end{equation}
where  $\lambda_1$, $\lambda_2$ and $\lambda_3$ are weights for each part of loss. $\phi(\theta)$ is the regularization term, e.g. $L_2$ norm. The definition of $L_{global}$ is same as $L_{local}$. To handle the severe class imbalance, we adopt weighted cross entropy loss \cite{Xie2015} for them:
\begin{equation}
L = -\frac{|Y_-|}{|Y_+|\gamma} \sum_{j \in Y_+}^{}\log P(y_j=1|X) - \sum_{j \in Y_-}^{}\log P(y_j=0|X) \\
\end{equation}
where $X$ is the input image, $y_j \in \{0, 1\}, j = 1, ..., |X|$ is the pixel-wise binary label map for $X$, $Y_+$ and $Y_-$ are the sets of positive and negative label pixels, $|Y_0|/(\gamma |Y_1|)$ is the weight for positive class, $\gamma$ is the hyperparameter to adjust the weight scale.

\textbf{Training strategy.} Training a deep neural network is a challenging task. For each input of LocalNet, only one patch of the output from GlobalNet is used, so the gradient of GlobalNet will be dominated by the patch easily, leading to unstable training. Apart from this, the available augmentations for fused network are not enough since doing fusion module requires pixel-wise aligning. This problem may lead to degradation of generalization. To address these issues, we first pre-train GlobalNet and LocalNet, and then freeze the layers before the fusion module, train the fusion module only until the network converges. Finally, we unfreeze all layers and fine-tune the whole network.

\section{Experimental Results}
\label{exper}

\subsection{Implementation Details}
Adam optimizer \cite{diederick2015} and polynomial decay \cite{yang2018} with an initial learning rate 0.0002 are used for both the training of GlobalNet and LocalNet. For fused net, we first train the fusion module for 10 epochs with learning rate 0.0002, then finetune the whole net with learning rate 1e-4 for 60 epochs, where Adam is used as the optimizer. The models are trained and tested with PyTorch \cite{paszke2017automatic} on two \textit{NVIDIA GeForce GTX1080}. It costs about 1 hour to train the GlobalNet, 4 hours to train the LocalNet and 4 to 8 hours to fine-tune the fused net. The inference of each $256^2 \times 3$ patch costs around 10 to 30 ms.

\begin{figure}[t!]
	\centering
	\includegraphics[width=9cm]{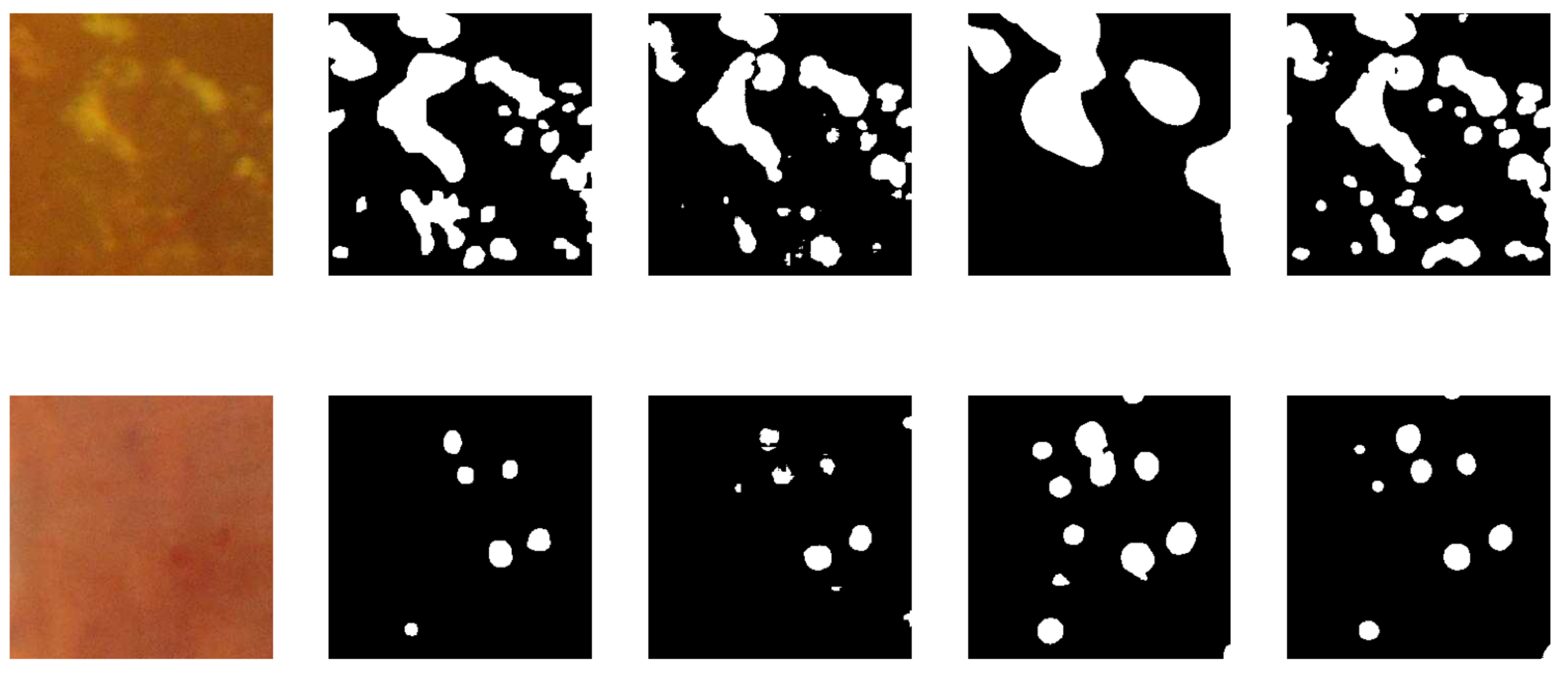}
	\caption{Example results of lesions segmentation. From left to right: input image patch, ground truth, segmentation result of LocalNet, GlobalNet and the proposed method. First row shows the example of lesion EX while second row shows the example of lesion MA.}
	\label{fig:example}
\end{figure}

\begin{figure}[t!]
	\centering
	\includegraphics[width=9cm]{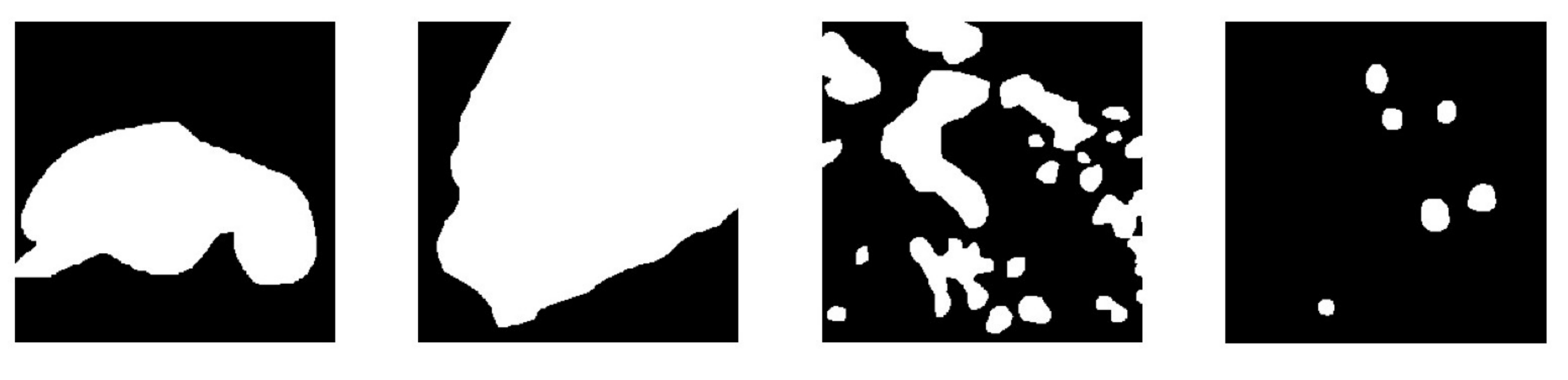}
	\caption{Examples of lesion patch labels. From left to right: HE, SE, EX, MA. The examples show that the areas of EX and MA are much smaller and more scattered than HE and SE.}
	\label{fig:example2}
\end{figure}

\subsection{Evaluation}
\textbf{Metrics.} In this paper, we utilize Area Under Precision Recall curve (AUPR) as our evaluation metric, which is same as the one used in the 2018 ISBI grand challenge. Precision (PPV) and True Positive Rate (TPR) are defined as follows:
\begin{equation}
	PPV = \frac{TP}{TP + FP}, \quad TPR = \frac{TP}{TP + FN}
\end{equation}
where true positives (TP) are lesion pixels that classified correctly, false positives (FP) are non-lesion pixels that incorrectly classified as lesion pixels, false negatives (FN) are lesion pixels that incorrectly classified as non-lesion. The precision-recall curve is obtained by plotting the precision-recall pairs given different thresholds, which are set as all the non-equal values of the lesion probability map.

\footnotetext[2]{https://idrid.grand-challenge.org/}

\textbf{Comparisons.} We compared the performance of the proposed fused network against two base models, LocalNet and GlobalNet mentioned above. The quantitative results are shown in Table \ref{tab:result1}, from which, we can see that our fused model outperforms the other two methods for EX and MA. However, for HE and SE, although our method is superior than LocalNet, still worse than GlobalNet. We discuss the reasons as below. For MA and EX, as seen in Fig \ref{fig:example}, the segmentation results of GlobalNet are very coarse due to the lost of details. The proposed fusion strategy can effectively compensate this drawback, achieving better results. However, as can be seen from Fig \ref{fig:example2}, the lesion areas of HE and SE are very large and compact, where the details that GlobalNet loses due to downsampling are ignorable. So in this case, GlobalNet can capture more useful features than LocalNet. Therefore, we can have a finding that the proposed network can improve the segmentation performance when the target lesions are scattered and of small size. We also refer the readers to the leaderboard of 2018 ISBI grand challenge \footnotemark[2] , where we are beyond all of the reported results. Even though our results maybe not significantly higher than the leaderboard, that is due to we use U-Net as our backbone, rather than duplicate their networks as backbones. We believe our framework can also work with other network backbones and improve their performances.

\begin{table}[t!]
	\centering
    \caption{Segmentation results of four type of lesions.}
	\begin{tabular}{ llllr }
		\firsthline
		Method/AUPR & EX & MA & HE & SE \\
		\hline
		\hline
		GlobalNet & 0.849  & 0.484 & \textbf{0.711}  & \textbf{0.720} \\
		LocalNet & 0.845 & 0.433 & 0.696 & 0.653 \\
		Our fused model & \textbf{0.889} & \textbf{0.525} & 0.703 & 0.679 \\
		\lasthline
	\end{tabular}
	\label{tab:result1}%
\end{table}

\section{Conclusion}
\label{sec:foot}
For segmenting small-size lesions in high-resolution retinal fundus images, downsampling-based methods will lose detailed information and patch-based methods are difficult to capture global contexts. Therefore, both of them may lead to performance degradation. In this paper, we proposed an end-to-end mutually local-global U-nets to solve this problem. The model consists of a global segmentation branch and a local(patch) segmentation branch, which are fused and jointly optimized, better capturing both the local details and the global contexts. The experimental results demonstrated the efficacy of our proposed method.

Since currently there is no large similar dataset, we plan to collect more data by ourselves and test the framework in future research. In addition, we believe the proposed fused model is not only applicable in retinal fundus lesions segmentation but also can be extended to other segmentation tasks.


\bibliographystyle{IEEEbib}
\bibliography{strings,refs}

\end{document}